\documentclass[sigplan,authorversion, nonacm, twocolumn]{acmart}
\usepackage{caption}
\usepackage{subcaption}

\usepackage{tikz}
\usetikzlibrary{arrows.meta}
\tikzset{%
	>={Latex[width=2mm,length=2mm]},
	base/.style = {rectangle, rounded corners, draw=black,
		minimum width=4cm, minimum height=1cm,
		text centered, font=\sffamily},
	task/.style = {base, fill=blue!20,minimum width=2.5cm},
	device/.style = {text=orange, minimum width=2.5cm,text opacity=1, fill opacity=0},
	phase/.style = {base, fill=orange!30,minimum width=2cm},
	application/.style = {base, fill=green!30},
	service/.style = {base, draw=blue},
}
\pgfdeclarelayer{bg1}    % declare background layer
\pgfdeclarelayer{bg2}    % declare background layer
\pgfsetlayers{bg1,bg2,main}  % set the order of the layers (main is the standard layer)

\usepackage[utf8]{inputenc}
\usepackage{hyperref}
\hypersetup{colorlinks=true, urlcolor= blue, citecolor=blue}
\usepackage{comment}
\usepackage{times} % DO NOT CHANGE THIS
\usepackage{helvet} % DO NOT CHANGE THIS
\usepackage{courier} % DO NOT CHANGE THIS
\usepackage{graphicx} % DO NOT CHANGE THIS
\usepackage{graphicx} % DO NOT CHANGE THIS
\usepackage{natbib} % DO NOT CHANGE THIS
\usepackage{caption} % DO NOT CHANGE THIS

\DeclareCaptionStyle{ruled}%
{labelfont=normalfont,labelsep=colon,strut=off}

\usepackage[normalem]{ulem}

\title{Unraveling the Hidden Environmental Impacts of AI Solutions for Environment\\ Life Cycle Assessment of AI Solutions}

\author{Anne-Laure Ligozat}
\affiliation{%
	\institution{Univ. Paris-Saclay, LIMSI, CNRS, ENSIIE}
	\city{Orsay}
	\country{France}}

\author{Julien Lef\`evre}
\affiliation{%
	\institution{Univ. Aix-Marseille, CNRS, Centrale Marseille}
	\city{Marseille}
	\country{France}}

\author{Aur\'elie Bugeau}
\affiliation{%
	\institution{Univ. Bordeaux, Bordeaux INP, CNRS, Laboratoire LaBRI}
	\city{Talence}
	\country{France}}

\author{Jacques Combaz }
\affiliation{%
	\institution{Universite Grenoble Alpes, VERIMAG}
	\city{Grenoble}
	\country{France}}

\begin{document}

\begin{abstract}
In the past ten years, artificial intelligence has encountered such dramatic progress that it is now seen as a tool of choice to solve environmental issues and in the first place greenhouse gas emissions (GHG). At the same time the deep learning community began to realize that training models with more and more parameters requires a lot of energy and as a consequence GHG emissions. To our knowledge, questioning the complete net environmental impacts of AI solutions for the environment (AI for Green), and not only GHG, has never been addressed directly. In this article, we propose to study the possible negative impacts of AI for Green. First, we review the different types of AI impacts, then we present the different methodologies used to assess those impacts, and show how to apply life cycle assessment to AI services. Finally, we discuss how to assess the environmental usefulness of a general AI service, and point out the limitations of existing work in AI for Green.
\end{abstract}

\maketitle

\pagestyle{plain}
\noindent

%%%%%%%%%%%%%%%%%%%%%%%%%%%%%%%%%%%%%%%%%%
\section{Introduction}
In the past few years, the AI community has begun to address the environmental impacts of deep learning programs: \cite{strubell_energy_2019} highlighted the impacts of training NLP models in terms of energy consumption and in terms of carbon footprint,  \cite{schwartz_green_2020} proposed the concept of Green AI, and the AI community created several tools to evaluate machine learning energy consumption~\cite{anthony_carbontracker_2020,henderson_towards_2020,lacoste_quantifying_2019,lannelongue_green_2020}.

These impacts are mainly expressed in terms of energy consumption and associated greenhouse gas (GHG) emissions. Yet, as we will discuss later, this energy consumption represents only a part of the complete environmental impacts of such methods. \cite{rapport_decla_montreal_en} for example states that "it is in terms of their indirect effects on the global digital sector that AI systems will have a major impact on the environment".  In the same spirit, \cite{walsh2020artificial} warns that "optimising actions for a restricted set of parameters (profit, job security, etc) without
consideration of these wider impacts can lead to consequences for others, including one's future self as well as future generations".

Evaluating the impacts of an AI service is not fundamentally different from doing it for another digital service. However, AI presents specificities, that must be taken into account because they increase its environmental impacts.

First, AI - and in particular deep learning - methods usually require large quantities of data. These data have to be acquired, transferred, stored and processed. All these steps require equipment and energy, and have environmental impacts. In the case of a surveillance satellite, the data will probably be in large quantities, but the number of acquisition devices may be limited; in the case of a smart building infrastructure, the data may be in smaller quantities, but many devices will be required. 

Training deep neural models also takes a lot of computation time and resources, partly because the model itself learns a comprehensive representation that enables it to better analyze the data. Whereas with other models, a human will provide part of this information, often in the form of a handcrafted solution. The computation cost can be even higher if the model does continuous learning.

At the same time, AI's popularity is increasing and AI is often presented as a solution to environmental problems with AI for Green proposals
~\cite{rolnick_tackling_2019,vinuesa_role_2020,gailhoferrole}. The negative environmental impacts can be briefly evoked - and in particular rebound effects %Jevons' paradox
\cite{rolnick_tackling_2019,Wu-2021} where unitary efficiency gains can lead to global GHG increase - but no quantification of all AI's environmental costs is proposed to close the loop between AI for Green and Green AI. That is why it is even more important to be able to assess the actual impacts, taking into account both positive and negative effects.

Incidentally those works often use the term AI to actually refer to deep learning methods, even though AI has a much wider scope with at least two major historical trends \cite{cardon2018neurons}. In this paper, we will also focus on deep learning methods, which pose specific environmental issues, and as we have seen, are often presented as possible solutions to environmental problems. 
%In this paper, 
We describe these impacts and discuss how to take them into account. 

Our contributions are the following:
\begin{itemize}
    \item We review the existing work to assess the environmental impacts of AI and show their limitations (Sections~ \ref{subsec:carbon} and \ref{subsec:AI4green}). 
    \item We present life cycle assessment (Section~\ref{subsec:lca}) and examine how it can comprehensively evaluate the direct environmental impacts of an AI service (Section~\ref{sec:estimate}).
    \item We discuss how to assess the environmental value of an AI service designed for environmental purposes (Section~ \ref{sec:5}).
    \item We argue that although improving the state of the art, the proposed methodology can only show the technical potential of a service, which may not fully realize in a real-life context (Section~\ref{sec-discussion}).
\end{itemize}

\section{Related work} \label{sec:related-work}
This section reviews existing tools for evaluating environmental impacts of AI as well as green applications of AI. It ends with an introduction to life cycle assessment, a well-founded methodology for environmental impact assessment but still not used for AI services.

\subsection{Carbon footprint of AI}
\label{subsec:carbon}
Strubell et al. \cite{strubell_energy_2019} has received much attention because it revealed a dramatic impact of NLP algorithms in the training phase: the authors found GHG emissions to be equivalent to 300 flights between New York and San Francisco. Premises of such an approach were already present in \cite{li_evaluating_2016} for CNN with less meaningful metrics (energy per image or power with no indications on the global duration).

In \cite{schwartz_green_2020} the authors observe a more general exponential evolution in deep learning architecture parameters. Therefore they promote "Green AI" to consider energy efficiency at the same level as accuracy in training models, and recommend in particular to report floating-point operations. Other authors \cite{garcia-martin_eva_estimation_2019} have also reviewed all the methods to estimate energy consumption from computer architecture. They distinguish between different levels of description, software/hardware level, instruction/application level and they consider how those methods can be applied to monitor training and inference phases in machine learning. 

In the continuity of \cite{strubell_energy_2019} and \cite{schwartz_green_2020}, several tools have been proposed to make the impacts of training models more visible. They can be schematically divided into
\begin{itemize}
\item \emph{Integrated tools}, such as Experiment Impact Tracker~\footnote{https://github.com/Breakend/experiment-impact-tracker} \cite{henderson_towards_2020}, Carbon Tracker~\footnote{https://github.com/lfwa/carbontracker} \cite{anthony_carbontracker_2020} and CodeCarbon~\footnote{https://codecarbon.io/}, which are all Python packages reporting measured energy consumption and the associated carbon footprint.
\item \emph{Online tools}, such as Green Algorithms \footnote{http://www.green-algorithms.org/} \cite{lannelongue_green_2020} and ML CO2 impact \footnote{https://mlco2.github.io/impact/\#compute} \cite{lacoste_quantifying_2019}, which require only a few parameters, such as the training duration, the material, the location but are less accurate.
\end{itemize}

AI literature mostly addresses a small part of direct impacts and neglects production and end of life, thus not following recommendations such as \cite{lca_standard_4_ict}. %These aspects will be detailed in the following section. 
In \cite{gupta2020chasing, Wu-2021} the authors point out the methodological gaps of the previous studies focusing on the use phase. In particular, manufacturing would account for about $75 \%$ of the total emissions of Apple or of an iPhone 5, just to give examples of various scales. Their study is based on a life cycle methodology, relying on sustainability reports with the GHG protocol standard. \cite{ml_practical_guide} provides a list of the carbon emission sources of an AI service, which gives a more comprehensive view of the direct impacts in terms of carbon footprint only. \cite{kaack:hal-03368037} also advocates the need for taking indirect impacts (e.g., behavioral or societal changes due to AI) into account when evaluating AI services. 

Some works focus on optimizing the AI processes regarding runtime, energy consumption, or carbon footprint. % developing more optimized methodologies...... 
For example, in \cite{patterson2021carbon} the authors update the results from \cite{strubell_energy_2019} and reveal a considerable reduction of the GHG impact - by a factor of 100 - if one considers the location of the data center used for training (low-carbon energy) and the architecture of the deep network (sparsity). Nevertheless as they  recognize, their study evaluates the GHG emissions of operating computers and data centers only and limits the perimeter by excluding the production and the end-of-life phases of the life cycle. Their work also considers a highly optimized use case, which may not be representative of real case scenarios. The energy efficiency of machine learning has also been the subject of dedicated workshops~\footnote{Workshop on Energy Efficient Machine Learning and Cognitive Computing, \url{https://www.emc2-ai.org/virtual-21}}.

\subsection{AI for Green benefits}
\label{subsec:AI4green}
When designing an AI for Green method i.e., a method using AI to reduce energy consumption or to benefit other environmental indicators, complete AI’s impacts should also be considered to build meaningful costs/benefits assessments. \cite{bommasani_opportunities_2021} proposes a framework for such cost-benefit analysis of AI foundation models to evaluate environmental
and societal trade-offs. We discuss this framework in Section~\ref{sec:5}. 
Most AI solutions for the environment lack a rigorous evaluation of the cost/benefit balance, and one of our contributions is to advance this issue.

\subsection{Life cycle assessment}
\label{subsec:lca}
LCA is a widely recognized methodology for environmental impact assessment, with ISO standards (ISO 14040 and 14044) and a specific methodology standard for ICT from ETSI/ITU~\cite{lca_standard_4_ict}.
It quantifies multiple environmental criteria and covers the different life cycle phases of a target system.
\cite{hauschild2018life} clearly states that "to avoid the often seen problem shifting where solutions to a problem creates several new and often ignored problems, these decisions must take a systems perspective. They must consider [...] the life cycle of the solution, and they need to consider all the relevant impacts caused by the solution."
The LCA theoretical approach exposed in \cite{heijungs2002computational} describes the system of interest as a collection of building blocks called \emph{unit processes}, for example "Use phase of the server" on which the model is trained. 
The set of all unit processes is called the \emph{technosphere}, as opposed to the \emph{ecosphere}. Each unit process can be expressed in terms of \emph{flows} of two kinds: 
\begin{itemize}
    \item \emph{Economic flows} are the directed links between the unit processes or said differently exchanges inside the technosphere.
    \item \emph{Ecnvironmental flows} are the links from the biosphere to the technosphere or vice versa.
\end{itemize}
The detailed description of such a system is called the life cycle inventory (LCI)~\footnote{We include the complete LCI of a generic AI service in the supplementary material to the paper.} and it can be formulated in terms of linear algebra. The goal of a life cycle assessment consists in computing the sum of the environmental flows of the system associated with a \emph{functional unit}. To be concrete, if one considers a heating system in a smart building, the functional unit could be "heating 1m$^2$ to 20$^{\circ}$C for one year".

Of course, very often, the LCI does not correspond exactly to the functional unit. The size of economic flows may not match (e.g., the functional unit may partially use shared servers and sensors), and a process may be \emph{multifunctional} , i.e., producing flows of different types at the same time (e.g., storage capacity and computational power). Both these problems can be solved using for instance \emph{allocation} methods according to a \emph{key}. A typical allocation key for network infrastructures would be the volume of data. For a data center it could be the economic value of storage and computational services when they cannot be physically isolated.

Even though LCA is widely used in many domains, it has rarely been applied to AI services.

\section{Life cycle assessment of an AI solution}

\label{sec:estimate}

When it comes to quantifying the impacts of digital technologies and in particular AI technologies, one faces several methodological choices that deserve a specific definition of the studied system. For instance, assessing the global impacts of the AI domain - if we could circumscribe it precisely - is not the same as assessing the impacts of an AI algorithm or service. 
The emerging field of AI's impacts quantification still suffers from a lack of common methodology, and in particular it very often focuses only on the Use phase of devices involved in an AI service. To perform meaningful quantification, we strongly suggest following the general framework of life cycle assessment (LCA, detailed in \ref{sec:lca}). We will show how it can be adapted to an AI service i.e., in this case a deep learning code used either alone or in a larger application.

AI being part of the Information and Communication Technology (ICT) sector, and following the taxonomies from \cite{hilty2010ict,Horner_2016}, its impacts can be divided into first-, second- and third-order impacts. In this section we focus only on first-order impacts while we will discuss second and third orders in Sections~\ref{sec:5} and \ref{sec-discussion}.

We will use the term \emph{AI service} for all the equipment (sensors, servers...) used by the AI, and the term \emph{AI solution} for the complete application using AI. In the case of the smart building, the \emph{AI solution} is the smart building itself, while the \emph{AI service} is the digital equipment needed for the smart infrastructure. 

\subsection{First-order impacts of an AI service}

\emph{First-order} - or \emph{direct} - impacts of the AI service are the impacts due to the different life cycle phases of the equipment:
\begin{itemize}
    \item \emph{Raw material extraction}, which encompasses all the industrial processes involved in the transformation from ore to metals;
    \item \emph{Manufacturing}, which includes the processes that create the equipment from the raw material;
    \item \emph{Transport}, which includes all transport processes involved, including product distribution; 
    \item \emph{Use}, which includes mostly the energy consumption of equipment while it is being used; 
    \item and \emph{End of life}, which refers to the processes to dismantle, recycle and/or dispose of the equipment.
\end{itemize}
For simplicity reasons, we will merge the first three items into a single \emph{production} phase in the rest of the paper.

For example an AI solution in a smart building may need sensors and servers that require resources and energy for their production, operation and end of life.

\begin{figure}
    \centering
    \includegraphics[width=.6\linewidth]{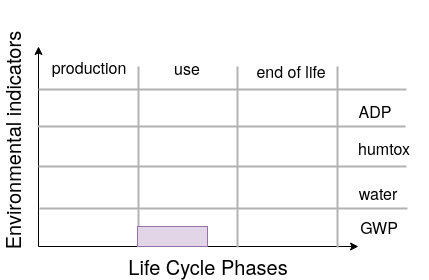}
    \caption{LCA dimensions: the first dimension corresponds to the phases of life cycle, the second one to the environmental impacts (see \ref{sec:estimate} for more details on this last dimension). }
    \label{fig:lca_dim}
\end{figure}

A second dimension is necessary to assess the impacts, a set of environmental criteria considered. Indeed each life cycle phase has impacts on different environmental indicators: Greenhouse Gases emissions (usually expressed as Global Warming Potential, GWP), water footprint, human toxicity, or abiotic resource depletion (ADP) for instance. 
In general, evaluating the environmental impact of a service requires multiple impacts criteria~\cite{lca_standard_4_ict}. ISO states that "the selection of impact categories shall reflect a comprehensive set of environmental issues related to the product system being studied, taking the goal and scope into consideration". Additionally, "the selection of impact categories, indicators and models shall be consistent with the goal and scope of the LCA study". Hence, 
the costs must take into account at least the criteria that are supposed to be tackled by the AI solution in the case of AI for Green: if the AI solution is applied to reduce energy consumption for example, the main expected gain will probably be in terms of carbon footprint, so at least the carbon footprint of using the model should be considered. For an application monitoring biodiversity, the most relevant criterion may be natural biotic resources (and not carbon footprint), which include wild animals, plants etc. 

Figure~\ref{fig:lca_dim} sums ups these two dimensions. As it has been previously stated, in the literature, only part of the global warming potential due to the use phase has generally been considered when evaluating AI, which corresponds to the shaded area in the figure.

\subsection{Life cycle assessment methodology for AI}
\label{sec:lca}
In this section, we focus on life cycle assessment of the AI solution, and the associated ICT equipment. We aim at proposing a methodology for applying the general framework of LCA to AI services. For LCA of all other processes, we refer to LCA standards and \cite{hauschild2018life} for example. 
In order to concretely apply the methodology presented for an AI service, we use the  ITU recommendation \cite{lca_standard_4_ict} for environmental evaluation of ICT.

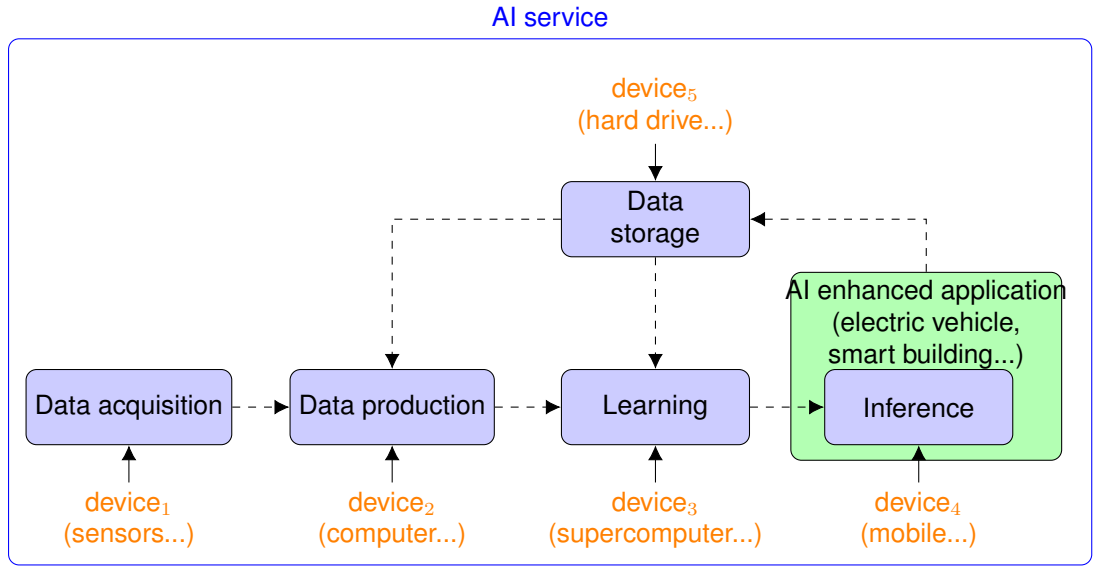
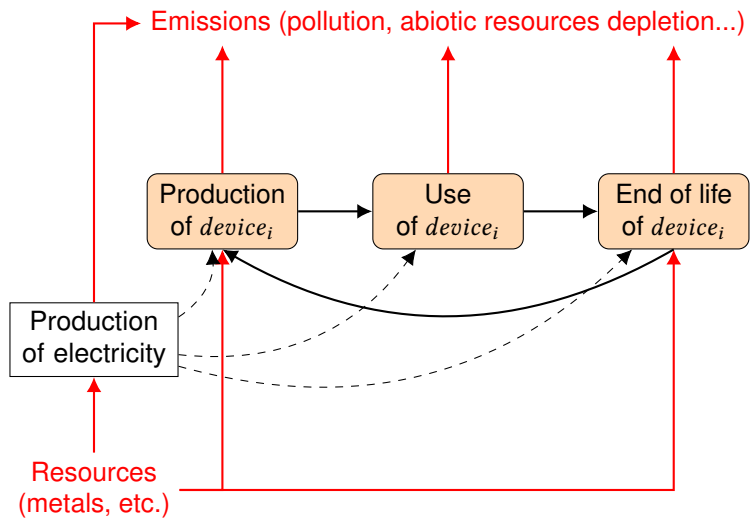
\begin{figure*}[h!]
    \centering
       \begin{subfigure}[]{.8\textwidth}
         \centering
            \begin{tikzpicture}[node distance=1.5cm,
    every node/.style={font=\sffamily}, align=center]
  \node (data_ac) [task] {Data acquisition};
  \node (data_prod) [task, right of=data_ac, xshift=2cm] {Data production};
  \node (learning) [task, text width=2cm, right of=data_prod, xshift=2cm] {Learning};
  \node (inference) [task, text width=2cm, right of=learning, xshift=2cm] {Inference};
  \node (data_stor) [task, text width=2cm, above of=learning, yshift=1cm] {Data storage};

  \node (device_ac) [device, below of=data_ac] {device$_1$\\(sensors...)};
  \node (device_prod) [device, below of=data_prod] {device$_2$\\(computer...)};
  \node (device_learn) [device, below of=learning] {device$_3$\\(supercomputer...)};
  \node (device_infer) [device, below of=inference] {device$_4$\\(mobile...)};
  \node (device_stor) [device, above of=data_stor] {device$_5$\\(hard drive...)};

  \begin{pgfonlayer}{bg2}
  \node (appli) at (10.6,1.8) [text centered, application, minimum height=2.5cm, minimum width=3.6cm, anchor=north] {}; 
  \node (appli_txt) at (10.6,1.1) [text opacity=1, fill opacity=0] {AI enhanced application\\(electric vehicle, \\smart building...)};
  \end{pgfonlayer}
  
  \begin{pgfonlayer}{bg1}
  \node (service) at (5.6,1.4) [service, minimum height=7cm, minimum width=14.4cm] {}; \node (service_txt) at (5.6,5.2) [text opacity=1, fill opacity=0, blue] {AI service};
  \end{pgfonlayer}

  \draw[->, dashed] (data_ac) -- (data_prod);
  \draw[->, dashed] (data_prod) -- (learning);
  \draw[->, dashed] (learning) -- (inference);
  \draw[->, dashed] (data_stor) -- (learning);
  \draw[->, dashed] (data_stor) -| (data_prod);
  \draw[->, dashed] (appli) |- (data_stor);

  \draw[->] (device_ac) -- (data_ac);
  \draw[->] (device_prod) -- (data_prod);
  \draw[->] (device_learn) -- (learning);
  \draw[->] (device_stor) -- (data_stor);
  \draw[->] (device_infer) -- (inference);

\end{tikzpicture}

         \caption{Different tasks involved in an AI service\\}
         \label{fig:schema_tasks}
     \end{subfigure}
     \bigskip
        \begin{subfigure}[]{\textwidth}
         \centering
             \begin{tikzpicture}[every node/.style={font=\sffamily}, align=center]
  \node (dev_prod) [phase] {Production\\of $device_i$};
  \node (dev_use) [phase, right of=dev_prod, xshift=2cm] {Use\\of $device_i$};
  \node (dev_eol) [phase,right of=dev_use, xshift=2cm] {End of life\\of $device_i$};

  \node (prod_elec) [draw,text width=2cm, below left of=dev_prod, yshift=-1cm, xshift=-1cm] {Production\\of electricity};
  \node (resources) [red, below of=prod_elec, yshift=-1cm] {Resources \\(metals, etc.)};

  \node (emissions) [text=red, above of=dev_use, yshift=1.5cm] {Emissions (pollution, abiotic resources depletion...)};

  \draw[->,red,thick] (dev_prod) -- (emissions.south -| dev_prod.north);
  \draw[->,red,thick] (dev_use) -- (emissions);
  \draw[->,red,thick] (dev_eol) -- (emissions.south -| dev_eol.north);

  \draw[->,red,thick] (resources) -| (dev_prod);
  \draw[->,red,thick] (resources) -| (dev_eol);

  \draw[->,red,thick] (resources) -- (prod_elec);
  
  \draw[->,red,thick] (prod_elec) |- (emissions);
  
  \draw[->,thick] (dev_prod) -- (dev_use);
  \draw[->,thick] (dev_use) -- (dev_eol);
  %\draw[->,thick] (dev_eol.south) -| ++(-.2cm,-1cm) -| (dev_prod.south);
  \draw[->,thick] (dev_eol.south) to[bend left] (dev_prod.south);

  \draw[->,dashed] (prod_elec) to[bend right] (dev_prod);
  \draw[->,dashed] (prod_elec) to[bend right] (dev_use);
  \draw[->,dashed] (prod_elec) to[bend right] (dev_eol);

\end{tikzpicture}
         \caption{Life cycle phases of each $device_i$ used by the service}
         \label{fig:schema_phases}
     \end{subfigure}
   
    \caption{Diagram representing the Life Cycle Inventory of an AI service: 
    Above: an AI for green application corresponds to the inference step that depends on other unit processes that require various devices. Below: the use of devices is located in a more global environment, including production of resources and impacts. In both schemes colored boxes correspond to unit processes, black arrows correspond to economic flows (bold: material, dashed: energy) and red arrows to environmental flows.
    }
    \label{fig:ai_lca}
\end{figure*}

Figure~\ref{fig:ai_lca} shows two sides of the Life cycle of an AI service. The top part of this figure shows the different tasks involved in an AI service, from a software point of view (data acquisition, ..., inference). For each task, one or several devices is used. The bottom part of the figure shows the life cycle phases of each of these devices, from a hardware point of view. The environmental impacts of the AI service will stem from the life cycle phases of the devices. Note that all devices involved in the AI tasks should be taken into account.

Remark on terminology: In the paper, the term "Use phase" refers to the use phase of the life cycle of equipment, corresponding to the devices provided for the AI service (box "Use of device" of the lower part in Figure~\ref{fig:ai_lca}). We call "Application phase" the inference phase of the AI service (green box of the upper part in Figure~\ref{fig:ai_lca}).

Concerning the system boundaries, we refer to \cite{ademe_digital_services_std} to consider the equipment for three tiers: 
\begin{itemize}
    \item \emph{terminals}. In the case of the smart building, this can include: user terminals used to develop, train and use the AI service; terminals in the facility where the AI service is trained dedicated to IT support; smart thermostats.
    \item \emph{network}. For the smart building case, network equipment used for training the AI model in the facility, and network equipment in the buildings where the thermostats are used.
    \item \emph{data center/server}. For the smart building case, servers on which the model is trained and used; training and inference can be done on the same server or not.
    
\end{itemize}
For each tier, all support equipment and activities  may also be considered. For example the power supply unit and HVAC of the data center should be taken into account. 

The life cycle stages to consider are the ones previously mentioned: production, use and end of life. 
In particular, \cite{lca_standard_4_ict} and \cite{ademe_digital_services_std} give classifications of unit processes according to the life cycle stages, which can be applied to AI services, as shown in Table~\ref{tab:itu4ai}. 
\setlength{\tabcolsep}{3pt}
\begin{table}[h!]
	\caption{Application to AI services of  ITU recommendation \cite{lca_standard_4_ict}  regarding the evaluation of life cycle stages/unit processes }
	\begin{tabular}{p{.1\linewidth}p{.58\linewidth}p{.25\linewidth}}
	Life cycle id & Life cycle stage and unit processes & Recommendation \\
	\hline
		\multicolumn{2}{l}{ A - Raw material acquisition } & Mandatory \\
	\hline
		\multicolumn{3}{l}{ B - Production }\\
		&Devices  production and assembly & Mandatory\\
		& Manufacturer support activities  &Recommended \\
		&Production of support equipment &Mandatory \\
		& ICT-specific site construction &Recommended \\
	\hline
		\multicolumn{3}{l}{C - Use}\\
		&Use of ICT equipment &Mandatory\\
		&Use of support equipment &Mandatory\\
		&Operator support activities & Recommended \\
		&Service provider support activities & Recommended \\
	\hline
		\multicolumn{3}{l}{D - End of life}\\
		& Preparation of ICT goods for reuse & Mandatory \\
		& Storage / disassembly / dismantling / crushing& Mandatory\\
	\end{tabular}
\label{tab:itu4ai}
\end{table}

If applied to our smart building use case, the unit processes that must be taken into account would be:
\begin{itemize}
    \item For equipment that is dedicated to the application, such as the smart thermostats:  Production, Use and End of life. 
    \item For the servers on which the AI service is trained and used and their environment (network devices, storage servers, backup servers, user terminal, HVAC... and other potential equipment not dedicated to the application): 
    \begin{itemize}
        \item Production and End of life with an allocation of the impacts, with respect to the execution time for instance.
        \item Part of the use phase corresponding to the dynamic energy consumption i.e., raise of consumption due to the execution of the program.
        \item Part of the use phase corresponding to the static consumption, with an allocation (for example if $n$ programs are run simultaneously, $1/n$ of this consumption) "since equipment is switched on in part to address the computing needs of the (Machine Learning) model" \cite{ml_practical_guide}.
    \end{itemize}
\end{itemize}

The production phase is generally important for ICT equipment in terms of global warming potential at least. Yet, when trying to assess this phase for deep learning methods, we are faced with a lack of LCAs for Graphical Processing Unit (GPUs) (or Tensor Processing Unit (TPUs) or equivalents). \cite{berthoud:hal-02549565}  yet showed that for a CPU-only data center in France, around 40\% of the GHG emissions of the equipment were due to the production phase.

The use phase is mostly due to the energy use, so the impacts of this part are highly dependent on the server/facility efficiency and the carbon intensity of the energy sources.

The end-of-life phase is difficult to assess in ICT in general because of lack of data concerning this phase of equipment. In particular, the end of life of many ICT equipment is poorly documented: globally, about 80\% of electronic and electrical equipment is not formally collected \cite{balde2017global}. \\

\section{Assessing the usefulness of an AI for Green service}
\label{sec:5}
Now that we have presented how the general framework of life cycle assessment can be adapted to AI solutions, we propose to use it for evaluating the complete benefits of an AI for Green service.

In this section, we will consider the following setting:
\begin{itemize}
    \item A reference application $M_1$ which corresponds to the application without AI. If the application is a smart building for example, $M_1$ will be the building without smart capabilities.
    \item An AI-enhanced application $M_2$ which corresponds to the application with an AI service that is supposed to have a positive impact on the environment. In the previous case, it would be the smart building.
\end{itemize}

\subsection{Theoretical aspects}

When proposing an AI for Green method, one should ensure that the overall environmental impact is positive: the positive gain induced by using the AI solution should be higher than the negative impacts associated to the solution.

\begin{figure}
    \centering
    \includegraphics[width=0.4\textwidth]{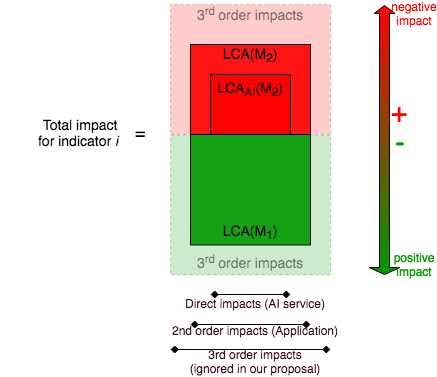}
    \caption{Overview of AI's impacts. First-order or direct impacts result from the equipment life cycle. Second-order impacts are the difference between the LCAs of the reference system and the AI-enhanced system. Third-order impacts are changes in technology or society induced by the application.} 
    \label{fig:ai_impacts}
\end{figure}

This requires to assess first-, second- and third-order impacts of AI~\cite{hilty2010ict,Horner_2016}, as illustrated in Figure~\ref{fig:ai_impacts}. As we detailed in the previous section, first-order impacts come from the life cycle phases of all the equipment necessary to develop and deploy the AI service.

\emph{Second-order} impacts correspond to the impacts due to the application of AI. AI can optimize or substitute existing systems: energy consumption in a building can be optimized using occupancy or behavior detection, energy profiling, etc.

\emph{Third-order} impacts are all changes in technology or society due to the introduction of AI solutions, possibly encompassing effects of very different scales, from individual behavioral responses to systemic and societal transformations, and from short-term to long-term effects. Rebounds effects fall into this category: an increase in efficiency does not necessarily translate into a reduction of impacts of the same magnitude, and it can even lead to an increase in these impact \cite{BERKHOUT2000}. Rebound effects occur because potential savings (in terms of money, time, resources, etc.) are transformed into more consumption \cite{Schneider2001}. For example, due to economical savings, smart building users may decide to increase heating temperature for better comfort or to buy more flight tickets after an increase in energy efficiency.

Third-order impacts are beyond the scope of the methodology proposed here, and are briefly discussed in Section~\ref{sec-discussion}. 
According to \cite{lca_standard_4_ict}, first and second-order impacts of the AI service should be estimated based on life cycle assessment (LCA), the difference between the two being the scope: for first-order impacts the scope is restricted to the equipment involved in the target AI service (for example the AI involved in a smart building), while second-order impacts consider the whole solution (the smart building itself). 
Including second-order impacts requires to extend the scope to the whole application AI is supposed to enhance. More specifically the net environmental impacts considering both first and second-order effects are obtained by computing:
\begin{equation}
    \Delta(M_2 | M_1) = LCA(M_2) - LCA(M_1) \in \mathbb{R}^d \label{eqDelta}
\end{equation}
with:
\begin{itemize}
    \item $M_1$ the reference application without using the AI service, 
    \item $M_2$ the application enhanced by AI,
    \item $LCA(x)$ a quantification of $d$ types of environmental impacts (e.g., GHG emissions, water footprint, etc.). LCA methodology is described in \ref{sec:lca}. Note that $LCA(M_2)$ includes the impacts of the AI service itself, i.e., $LCA_{AI}(M_2)$.
\end{itemize}

A previous work~
\cite{bommasani_opportunities_2021} also gave a simplified scheme for assessing the cost-benefit of deploying a foundation model, which also includes social benefits and costs, but does not explicit the direct environmental costs of using this model. We propose to relate our methodology (Equation \eqref{eqDelta}) to their proposal. 
%Propose a general scheme with an algorithm + life cycle phase including what could be its benefits 
Adopting their equation, but focusing on the environmental impacts only,
%and extending their reach\remab{en lisant rapidement, je n'ai pas l'impression que ce soit une extension, mais plutôt une simplification.}, 
the overall value of a model can be assessed with:
\begin{equation}
    V(M) = S(M) - E(M) - O(M)
    \label{eq1}
\end{equation}
%\remjl{j'ai repris les notations d'origine pour ne pas perdre le lecteur qui connaitrait} 
with:
\begin{itemize}
    \item $V(M)$ the value of using the model i.e., the environmental gain induced by its use in the practical application considered
    \item $S(M)$ the environmental benefit that can be interpreted as the difference between the initial impact of the application and its final impact (not taking into account the AI solution i.e., the Learning and Inference task in the top part of Figure \ref{fig:ai_lca}) 
    \item $E(M)$ the energy cost of the model % TODO : of the use phase? 
    \item $O(M)$ all other impacts including chip production, waste, risks for biodiversity, and third-order impacts (which are not discussed here).%the indirect (or second and third order) effects, including chip production, waste, risks for biodiversity
\end{itemize}
Regarding the well-established framework of LCA, this approach suffers from several weaknesses. First, in the equation all the values are expressed in dollars. This formally allows to perform addition of several kinds of impacts but with an arbitrary consideration to the diversity of environmental issues. By definition, LCA considers multiple criteria for the impacts, previously described at the beginning of Section \ref{sec:estimate} (GHG emissions, water footprint...). LCA may aggregate several impacts but with specific weights not necessarily dependent on an economic value. %\remjl{là, c'est sans ref et possiblement faux} \remal{nope c'est vrai. cf chapitre 10 de \cite{hauschild2018life} "Life Cycle Impact Assessment" sur la "normalization, weighting and aggregation", p.193 en particulier}.
As noted in \cite{hauschild2018life} "there is no scientific basis on which to reduce the
results of an LCA to a single result or score because of the underlying ethical
value-choices".

Besides, if one considers for instance the case of an AI service dedicated to biodiversity (see for instance 8.1 in \cite{rolnick_tackling_2019}), one would expect to precisely quantify the positive impact of this service on biodiversity (schematically, how many species can be saved?) balanced by the negative ones (producing chips for GPUs has an impact on the biodiversity through several sources of pollution \cite{villard2015drawing}). Adopting Equation \eqref{eq1} will mix several impacts together and may dilute the value of interest (e.g., biodiversity), that could be burdened by negative impacts regarding energy to train the models, for instance. 

Last, even if the equation is not wrong per se, the expression in terms of benefit/costs is questionable, and practical means for its computation are missing in \cite{bommasani_opportunities_2021}. 

We thus believe that Equation \eqref{eqDelta} should be used. 
Terms of Equation \eqref{eq1} can be related to the methodology proposed in our paper as follows:
\begin{equation}
     \underbrace{V(M_2)}_{- \Delta(M_2 | M_1) } \approx S(M_2|M_1) \underbrace{- E(M_2) - O(M_2)}_{LCA_{AI}(M_2)}
\end{equation}
where $\Delta(M_2 | M_1)$ and $LCA_{AI}(M_2)$ are defined in Equation \eqref{eqDelta}. The negative impacts of an AI solution $M_2$ compared to the reference solution $M_1$ are not always restricted to its AI part (i.e., to $E(M_2)$ and $O(M_2)$). For example, compared to a standard vehicle the negative impacts of an autonomous vehicle are not only due to the life cycle of (additional) ICT equipment, but also to additional aerodynamic drag due to the presence of LIDAR on the roof \cite{Taiebat2018}. Hence, the nature of the impacts in $S(M_2|M_1)$ (positive or negative) cannot be stated a priori and depends on complete LCA results for both applications $M_2$ and $M_1$. It may also depend on the target environmental criteria.

\subsection{Case studies}
In order to review the kind of evaluation that is usually made in the AI for Green literature, we analyzed the references for several domains of \cite{rolnick_tackling_2019}, which identifies potential applications of machine learning for climate change adaptation of mitigation~\footnote{This review was documented in a csv file, which is given as supplementary material to the paper.}. 

We mostly chose domains that had been flagged as having a \emph{High Leverage} and noted for each paper cited in the corresponding section the kind of environmental evaluation, with the following categories:
\renewcommand{\theenumi}{\alph{enumi}}
\begin{enumerate}
    \item No mention of the environmental gain.
    \item General mention of the environmental gain.
    \item A few words about the environmental gain but no quantitative evaluation or only indirect estimation.
    \item Evaluation of the energy gain without taking the AI service into account.
    \item Evaluation of the energy gain taking the use phase of the AI service into account.
    \item Comprehensive evaluation of the environmental gain (comparison of LCAs).
\end{enumerate}

\begin{figure*}[h!]
    \centering
    \includegraphics[width=.9\linewidth]{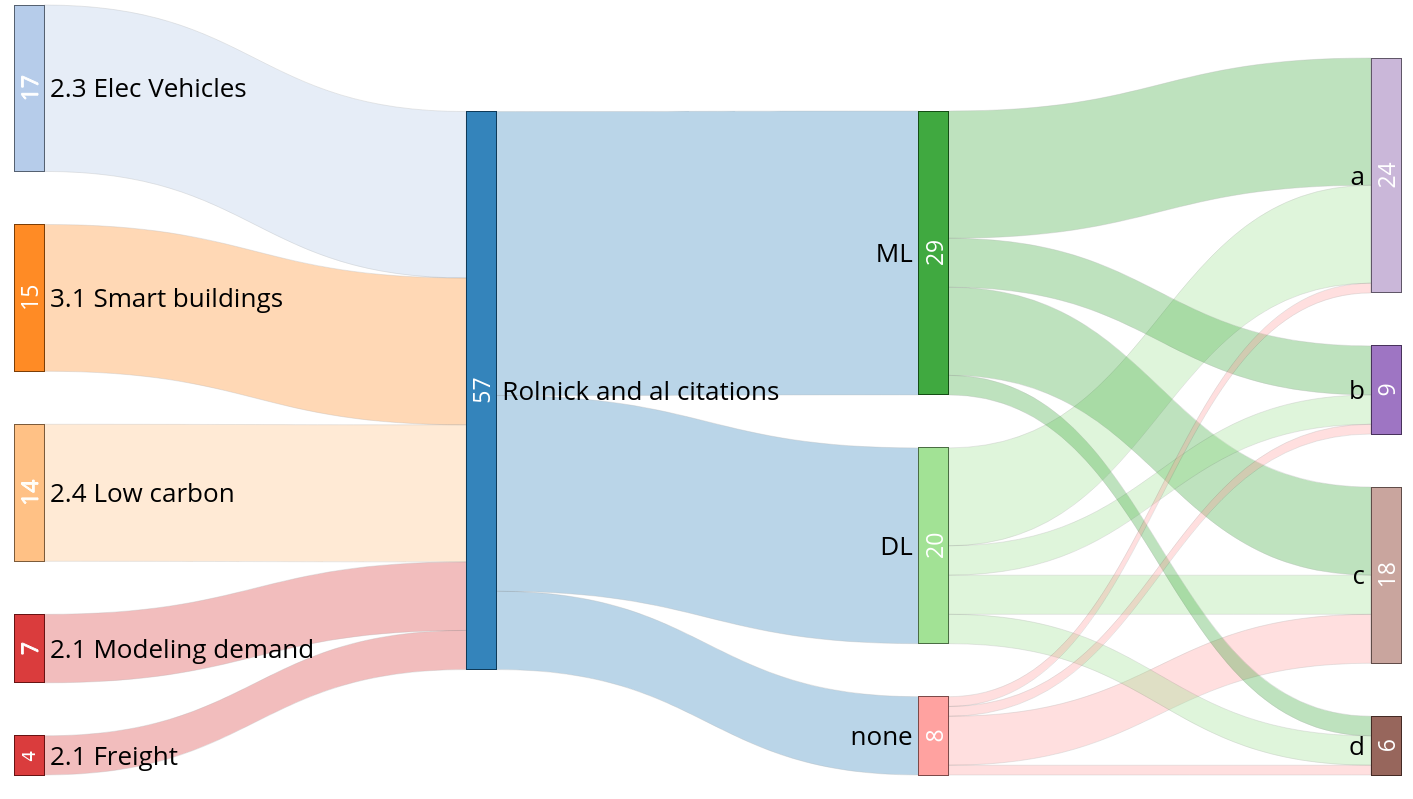}
    \caption{Sankey diagram of parts of Rolnick's paper references in terms of environmental evaluation \emph{{\small (created with the Sankey Diagram Generator by Dénes Csala, based on the Sankey plugin for D3 by Mike Bostock; https://sankey.csaladen.es; 2014)}}}
    \label{fig:rolnick_refs}
\end{figure*}

The results of the review are shown in Figure~\ref{fig:rolnick_refs}.

The central node is "Rolnick et al. citations". On its left are the domains of the citations. For example, the Smart building section contained 15 relevant citations. 

On its right the first flows show the partition into general machine learning applications (ML), deep learning applications (DL), and other methods (other). For example, 20 papers corresponded to deep learning applications.

The last flows on the right show the kinds of environmental evaluation. We can note that about half of the papers do not include any environmental evaluation, although the focus is on applications to tackle climate change. Many papers also give a distant proxy for evaluation, such as detailing the possible impacts without quantification, or indicating the execution time of the program.

A few citations evaluate the environmental gain, mostly in terms of energy gain, but none of the papers considered took into account the AI service impacts.

It can be noted that other papers that include an evaluation of part of these impacts, can be found in the literature. 
\cite{BRACQUENE2020232} for example present an intelligent control systems that takes into account the expected occupancy in order to adapt the thermostat and save energy. They do not take into account learning the occupancy model, but take into account the LCA of the smart thermostats, and show that the energy needed for these devices across their whole life cycle will almost always be lower that the energy saved.

\section{Discussion}\label{sec-discussion}

In this paper, we have analyzed the environmental impacts of AI solutions, in particular in the case of AI for Green applications, and proposed a framework to evaluate them more completely. The proposed methodology compares, through life cycle assessment, the impact of a reference solution with the AI one~\eqref{eqDelta} for the appropriate types of environmental impacts. The analysis of literature on AI solutions has made salient the following issues/problems.

\subsection{Current environmental evaluation of AI services is under-estimated}
We have shown that AI for Green papers only take into account a small part of the direct environmental impacts. 

Several reasons can explain this under-estimation. The narratives about dematerialization that would correspond to a dramatic decrease in environmental impacts, permeate AI as a part of ICT \cite{bol2021moore}.  However, these narratives have proven to be false until now. Attention to AI's GHG emissions has focused on electricity consumption (energy flows). At the moment, material flows receive less attention in AI. However, it is beginning to be considered \cite{gupta2020chasing,Wu-2021}.

\subsection{AI research should use Life Cycle Assessment to assess the usefulness of an AI service}
Life cycle assessment is a solid methodology to evaluate not only global warming potential but also other direct environmental impacts. LCA considers all the steps from production to use and end of life. However, it has several well-known limitations due to the complexity of processes involved in material production. Obtaining all the information to assign reliable values to each edge of the life cycle inventory also proves difficult, e.g., there is very little information on manufacturing impacts of GPU either from manufacturers or in LCA databases. To solve this problem, we could encourage the AI community to lobby companies to open a part of their data. This approach would be in the same spirit as what is happening for open science but would also require taking  legal issues into account.

\subsection{AI for Green gains are only potential}
Even when a properly conducted LCA concludes that an AI solution is environmentally beneficial, such a result should be considered with caution. Environmental benefits computed by the LCA-based methodology proposed in this paper correspond to a technical and simplistic view of environmental problems: it assumes that AI will enhance or replace existing applications, all other things being equal. The ambition to solve societal problems using AI is praiseworthy, but it should probably be accompanied by socio-technical concerns and an evaluation of possible third-order effects. For example, autonomous vehicles are often associated with potential efficiency gains (such as helping car sharing, or allowing platooning) and corresponding environmental benefits \cite{Taiebat2018}. However, autonomy could also profoundly transform mobility in a non-ecological way \cite{Coroama2020}.

\subsection{AI services and large deployment} Evaluating third-order effects is even more critical when large-scale deployment of the proposed solution(s) is projected, e.g., to maximize absolute gains. This case requires special attention even in LCA, since large-scale deployment may induce societal reorganizations for producing and operating the solution(s). For example, the generalization of AI may lead to a substantial increase in demand for specific materials (such as lithium or cobalt) or energy. This increase may have non-linear environmental consequences, e.g., opening new and less performing mines, increasing the use of fossil fuel based power plants, etc. Hence in this case, the \emph{attributional} LCA framework we suggest using in this paper needs to be replaced by the much more complex \emph{consequential} one \cite{hauschild2018life}.

\section*{Authors contribution}
``Conceptualization, all authors; methodology, all authors; validation, all authors; formal analysis, all authors; investigation, J.L. and A.-L. L.; data curation, J.L. and A.-L. L.; writing---original draft preparation, all authors; writing---review and editing, all authors; visualization, A.B. and A.-L. L.; supervision, A.-L. L.; project administration, A.-L. L. All authors have read and agreed to the published version of the manuscript.\\

\section*{Acknowledgments}This work was partly supported by the CNRS EcoInfo group\\
(\href{https://ecoinfo.cnrs.fr/}{https://ecoinfo.cnrs.fr/}).

%%%%%%%%%%%%%%%%%%%%%%%%%%%%%%%%%%%%%%%%%%
%% Optional
\section*{Abbreviations}{
The following abbreviations are used in this manuscript:\\

\noindent 
\begin{tabular}{@{}ll}
AI & Artificial Intelligence\\
CNN & Convolutional Neural Network\\
CPU & Central Processing Unit\\
DL & Deep Learning \\
GHG & Greenhouse Gas\\
GPU & Graphics Processing Unit\\
HVAC & Heating, Ventilation, and Air Conditioning\\
ICT & Information and Communications Technology\\
LCA & Life cycle Assessment or Analysis\\
LCI & Life Cycle Inventory\\
ML & Machine Learning \\
NLP & Natural Language Processing\\
TPU & Tensor Processing Unit
\end{tabular}}

%\begin{thebibliography}{}

%\end{thebibliography}

\bibliographystyle{apalike}
\end{document}